\pdfoutput=1
\documentclass[10pt,twocolumn,letterpaper]{article}

\usepackage[pagenumbers]{cvpr} 

\usepackage{graphicx}
\usepackage{amsmath}
\usepackage{amssymb}
\usepackage{booktabs}

\usepackage{rotating}
\usepackage{array}
\usepackage{colortbl}
\usepackage{multirow}
\usepackage{color}
\usepackage{enumitem}
\usepackage[dvipsnames]{xcolor}
\DeclareMathOperator*{\diag}{diag}
\usepackage{bbm}
\usepackage{arydshln}

\usepackage[pagebackref,breaklinks,colorlinks,citecolor=RoyalBlue]{hyperref}

\usepackage[capitalize]{cleveref}
\crefname{section}{Sec.}{Secs.}
\Crefname{section}{Section}{Sections}
\Crefname{table}{Table}{Tables}
\crefname{table}{Tab.}{Tabs.}

\def\cvprPaperID{532} 
\def\confName{CVPR}
\def\confYear{2023}

\begin{document}
	
	\title{SIM: Semantic-aware Instance Mask Generation for\\ Box-Supervised Instance Segmentation}
	
	\author{Ruihuang Li\thanks{denotes the equal contribution, \dag denotes the corresponding author. This work is supported by the Hong Kong RGC RIF grant (R5001-18).},\quad Chenhang He\footnotemark[1],\quad Yabin Zhang,\quad Shuai Li,\quad Liyi Chen,\quad Lei Zhang\footnotemark[2]\\
	The Hong Kong Polytechnic
	University\\
	{\tt\small \{csrhli, csche, cslzhang\}@comp.polyu.edu.hk} }
   
\maketitle

\begin{abstract}
	Weakly supervised instance segmentation using only bounding box annotations has recently attracted much research attention. Most of the current efforts leverage low-level image features as extra supervision without explicitly exploiting the high-level semantic information of the objects, which will become ineffective when the foreground objects have similar appearances to the background or other objects nearby. We propose a new box-supervised instance segmentation approach by developing a Semantic-aware Instance Mask (SIM) generation paradigm. Instead of heavily relying on local pair-wise affinities among neighboring pixels, we construct a group of category-wise feature centroids as prototypes to identify foreground objects and assign them semantic-level pseudo labels. Considering that the semantic-aware prototypes cannot distinguish different instances of the same semantics, we propose a self-correction mechanism to rectify the falsely activated regions while enhancing the correct ones. Furthermore, to handle the occlusions between objects, we tailor the Copy-Paste operation for the weakly-supervised instance segmentation task to augment challenging training data. Extensive experimental results demonstrate the superiority of our proposed SIM approach over other state-of-the-art methods. The source code: \url{https://github.com/lslrh/SIM}.             
\end{abstract}

\section{Introduction}
\label{sec:intro}
Instance segmentation is among the fundamental tasks of computer vision, with many applications in autonomous driving, image editing, human-computer interaction, \etc. The performance of instance segmentation has been improved significantly along with the advances in deep learning~\cite{chen2019hybrid,he2017mask,tian2020conditional,wang2020solov2}. However, training robust segmentation networks requires a large number of data with pixel-wise annotations, which consumes intensive human labor and resources. To reduce the reliance on dense annotations, weakly-supervised instance segmentation based on cheap supervisions, such as bounding boxes~\cite{tian2021boxinst,lee2021bbam,hsu2019weakly}, points~\cite{cheng2022pointly} and image-level labels~\cite{kim2022beyond,ahn2019weakly}, has recently attracted increasing research attention. 

\begin{figure}
	\centering 
	\includegraphics[scale=0.185]{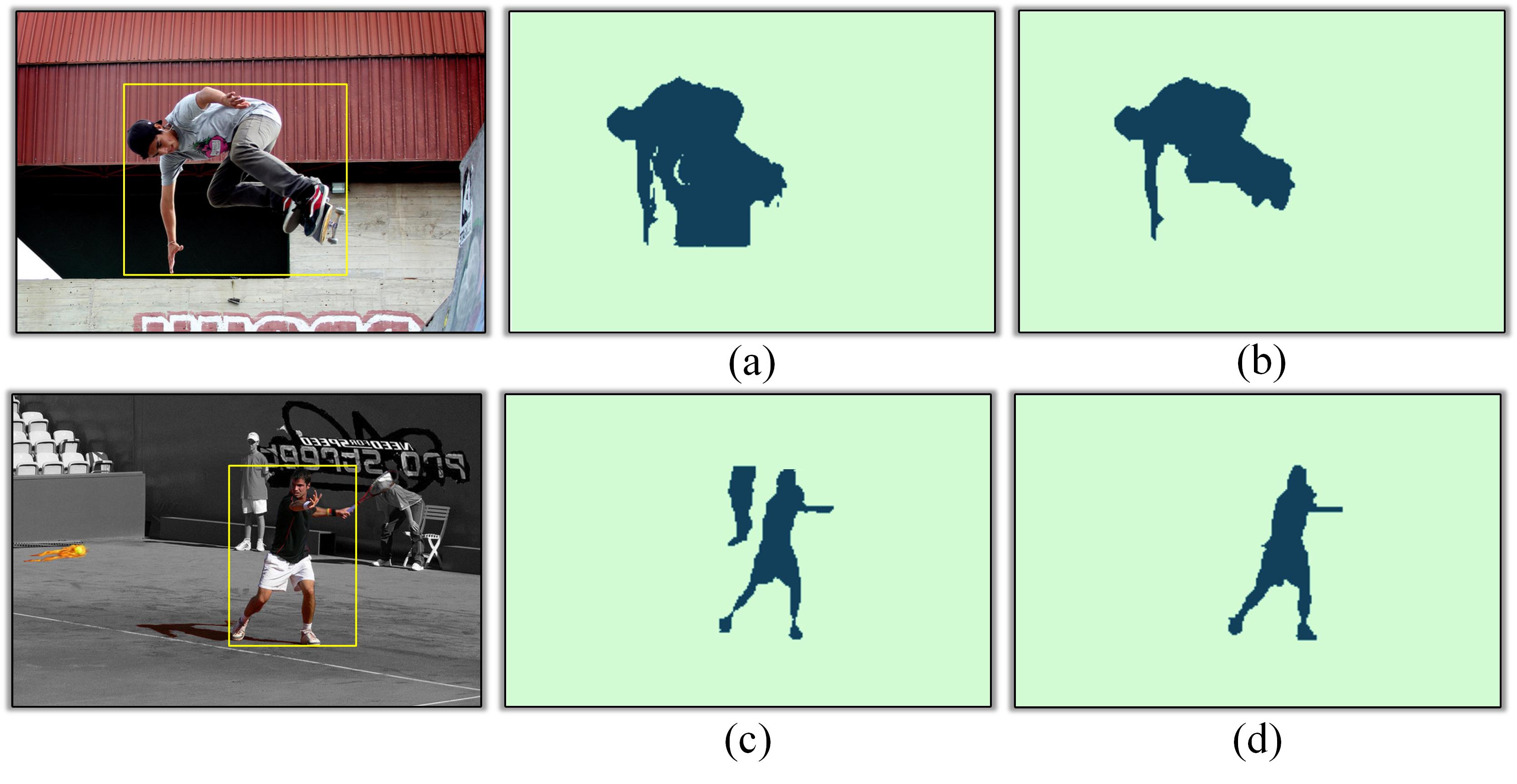}\\
	\vspace{-1em}
	\caption{The pipeline of Semantic-aware Instance Mask (SIM) generation method. (a) shows the mask prediction produced by using only low-level affinity supervision, where the foreground heavily blends with background. (b) and (c) show the semantic-aware masks obtained with our constructed prototypes, which perceive the entity of objects but are unable to separate different instances of the same semantics. (d) shows the final instance pseudo mask rectified by our proposed self-correction module.}
	\label{fig1}
\end{figure}

In this paper, we focus on box-supervised instance segmentation (BSIS), where the bounding boxes provide coarse supervised information for pixel-wise prediction task. To provide pixel-wise supervision, conventional methods~\cite{dai2015boxsup,kulharia2020box2seg} usually leverage off-the-shelf proposal techniques, such as MCG~\cite{pont2016multiscale} and GrabCut~\cite{rother2004grabcut}, to create pseudo instance masks. However, the training pipelines of these methods with multiple iterative steps are cumbersome. Several recent works~\cite{tian2021boxinst,hsu2019weakly} enable end-to-end training by taking pairwise affinities among pixels as extra supervision. Though these methods have achieved promising performance, they heavily depend on low-level image features, such as color pairs~\cite{tian2021boxinst}, and simply assume that the proximal pixels with similar colors are likely to have the same label. This leads to confusion when foreground objects have similar appearances to the background or other objects nearby, as shown in Fig.~\ref{fig1} (a). It is thus error-prone to use only low-level image cues for supervision since they are weak to represent the inherent structure of objects. 

Motivated by the fact that high-level semantic information can reveal intrinsic properties of object instances and hence provide effective supervision for segmentation model training, we propose a novel Semantic-aware Instance Mask generation method, namely SIM, to explicitly exploit the semantic information of objects. To distinguish proximal pixels with similar color but different semantics (please refer to Fig.~\ref{fig1} (a)), we construct a group of representative dataset-level prototypes, \ie, the feature centroids of different classes, to perform foreground/background segmentation, producing semantic-aware pseudo masks (see Fig.~\ref{fig1} (b)). These prototypes abstracted from massive training data can capture the structural information of objects, enabling more comprehensive semantic pattern understanding, which is complementary to affinity supervision of pairwise neighboring pixels. However, as shown in Fig.~\ref{fig1} (c), these prototypes are unable to separate the instances of the same semantics, especially for overlapping objects. We consequently develop a self-correction mechanism to rectify the false positives while enhancing the confidence of true-positive foreground objects, resulting in more precise instance-aware pseudo masks, as shown in Fig.~\ref{fig1} (d).

It is worth mentioning that our generated pseudo masks could co-evolve with the segmentation model without cumbersome iterative training procedures in previous methods~\cite{dai2015boxsup,lee2021bbam}. In addition, considering that the existing weakly-supervised instance segmentation methods only provide very limited supervision for rare categories and overlapping objects due to the lack of ground truth masks, we propose an online weakly-supervised Copy-Paste approach to create a combinatorial number of augmented training samples. Overall, the major contributions of this work can be summarized as follows:
\begin{itemize}[leftmargin=*]
	\item[$\bullet$] A novel BSIS framework is presented by developing a semantic-aware instance mask generation mechanism. Specifically, we construct a group of representative prototypes to explore the intrinsic properties of object instances and identify complete entities, which produces more reliable supervision than low-level features.  
	\vspace{-0.5em}
	\item[$\bullet$] A self-correction module is designed to rectify the semantic-aware pseudo masks to be instance-aware. The falsely activated regions will be reduced, and the correct ones will be boosted, enabling more stable training and progressively improving the segmentation results.    
	\vspace{-0.5em}
	\item[$\bullet$] We tailor the Copy-Paste operation for weakly-supervised segmentation tasks in order to create more occlusion patterns and more challenging training data. The overall framework can be trained in an end-to-end manner. Extensive experiments demonstrate the superiority of our method over other state-of-the-art methods.
\end{itemize}

\begin{figure*}
	\centering 
	\includegraphics[scale=0.42]{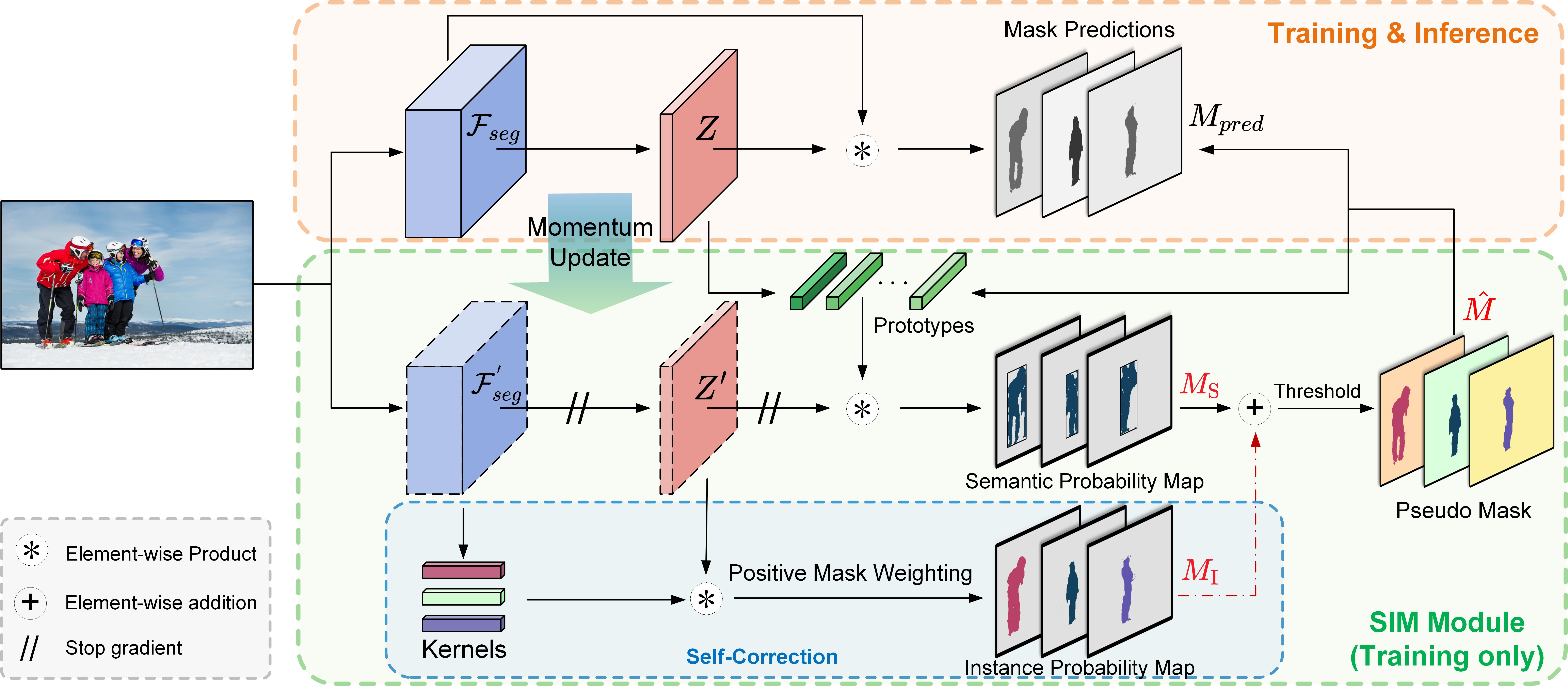}\\
	\caption{The framework of our proposed Semantic-aware Instance Mask (SIM) generation method. The model contains the main segmentation network $\mathcal{F}_{seg}$ and its momentum-updated version $\mathcal{F}'_{seg}$. Given an image $X$, we first pass it through $\mathcal{F}_{seg}$ and $\mathcal{F}'_{seg}$ to obtain the corresponding mask features $Z$ and $Z'$. The prototypes are then updated as the moving average of feature cluster centroids. Next, we obtain the semantic probability map $M_{\rm{S}}$ by measuring the distance between prototypes and mask features $Z'$. After that, the falsely activated instances in $M_{\rm{S}}$ are rectified by the instance probability map $M_{\rm I}$, which is obtained by integrating different positive masks of the same ground truth object. Finally, we obtain the pseudo mask $\hat{M}$ by selecting highly-confident pixels with two thresholds. }
	\label{fig2}
	\vspace{-1.0em}
\end{figure*}      
\section{Related Work}
\noindent{\bf Instance Segmentation (IS)} is a fundamental task in computer vision fields, which aims to predict the pixel-wise mask for each instance of interest in an image. Many top performing IS methods~\cite{liu2018path,huang2019mask,zhang2021refinemask,chen2019hybrid} follow the Mask R-CNN meta-architecture~\cite{he2017mask}, which splits the IS task into two consecutive stages and performs segmentation on the extracted region proposals. Single-stage IS methods have also been rapidly developed during the past few years. YOLACT~\cite{bolya2019yolact} and BlendMask~\cite{chen2020blendmask} employ fine-grained FPN features rather than the RoI-aligned features for mask prediction. However, they still need crop operation for object localization. Some methods segment each instance in a fully convolutional manner without resorting to the detection results. For example, CondInst~\cite{tian2020conditional} and SOLO~\cite{wang2020solov2} employ instance-aware conditional convolutions and dynamically generate convolution kernels to segment different objects. Universal architectures~\cite{Cheng_2022_CVPR,yu2022k} have emerged with DETR~\cite{carion2020end} and show that end-to-end set prediction architecture is general enough for any segmentation task. Despite the promising performance, these methods heavily rely on expensive pixel-wise mask annotation, which restricts their usability in many practical applications.  

\noindent{\bf Weakly-Supervised Instance Segmentation (WSIS)} with weak annotations is a more attractive yet challenging task. Some works attempt to achieve high-quality segmentation with box-level annotations~\cite{tian2021boxinst,hsu2019weakly,lee2021bbam, khoreva2017simple} or image-level annotations~\cite{kim2022beyond,ahn2019weakly}. Khoreva~\etal~\cite{khoreva2017simple} employ box supervisory training data for WSIS. However, the proposed method relies on the region proposal techniques, such as GrabCut~\cite{rother2004grabcut} and MCG~\cite{pont2016multiscale}, to generate pseudo masks in an offline manner. Other recent methods~\cite{lee2021bbam,wang2021weakly} also focus on generating instance labels by using an independent network, which require either extra salient data~\cite{wang2020solov2} or some post-processing methods~\cite{lee2021bbam}. This inevitably leads to a complicated training pipeline. 

To achieve a simple yet effective training pipeline, BBTP~\cite{hsu2019weakly} formulates WSIS as a multiple-instance learning problem and introduces a structural constraint to maintain the unity of estimated masks. BoxInst~\cite{tian2021boxinst} builds upon an efficient CondInst~\cite{tian2020conditional} framework, and enforces the proximal pixels with similar colors to have the same label through a pairwise loss. Despite the promising performance, these methods depend heavily on local color supervision while neglecting the global structure of the entire object. Different from these methods, our proposed method provides more reliable supervision by leveraging high-level semantic information, which is beneficial for capturing the intrinsic structures of objects.           

\noindent{\bf Pseudo Mask Generation.} A widely adopted technique in conventional weakly-supervised semantic segmentation methods is Class Activation Map (CAM)~\cite{zhou2016learning}, which aims to obtain an object localization map from class labels. However, CAM only identifies the most discriminative object regions and suffers from the problem of limited activation area~\cite{ahn2018learning,hou2018self,huang2018weakly,shimoda2019self,he2022voxel}. Given that bounding boxes could provide the location information of objects in an image, BBAM~\cite{lee2021bbam} employs an object detector to produce a bounding box attribute map, which serves as a pseudo ground truth mask. As a more lightweight approach, self-training-based methods~\cite{zou2018unsupervised,zou2019confidence, li2022class, zhang2021prototypical,li2021t} select high-scoring predictions on unlabeled data as pseudo labels for training. The idea of assigning labels based on prototypes has also been explored in semantic segmentation~\cite{zhou2022rethinking, zhou2022regional,li2022class}. In this work, the prototype technique is adapted to capture the global structure of objects with the same semantics, reducing the noise caused by low-level feature supervision.

\section{Method}
\subsection{Overview} 
\label{sec1}
In the setting of box-supervised instance segmentation (BSIS), we are given a set of box annotated training data $\mathcal{D}=\{X_n, Y_n, B_n\}^N_{n=1}$, where $N$ is the number of images. Besides, $Y_n=\{{ y}^k_n\}^{K}_{k=1}$ and $B_n=\{{ b}^k_n\}^{K}_{k=1}$ denote the class-level and box-level annotations, where $K$ is the number of instances in the image $X_n$, ${y}^k_n\in \{1,\cdots, C\}$ represents the category label of the $k$-th object in the $n$-th image, and ${b}^k_n$ specifies its corresponding location. 

The overview of our method is shown in Fig.~\ref{fig2}, where the proposed SIM module is highlighted in the green dotted box. We choose CondInst~\cite{tian2020conditional} and Mask2Former~\cite{Cheng_2022_CVPR} as the basic segmentation networks due to their simplicity and effectiveness. Instead of only relying on local pair-wise affinities among pixels as supervision~\cite{tian2021boxinst,hsu2019weakly}, we employ a group of semantic-level prototypes to capture global structural information of objects, and produce semantic probability map $M_{\rm S}$ by computing the distances between each pixel-wise feature vector and all prototypes. Since these prototypes are unable to separate different objects of the same semantics, we propose a self-correction mechanism to deactivate falsely estimated objects by using an instance probability map $M_{\rm I}$. This map can be obtained by integrating different positive masks corresponding to the same instance with an IoU-based weighting strategy. Finally, we employ two thresholds to select confident predictions as pseudo ground truths $\hat{M}$, and use them for training the segmentation network $\mathcal{F}_{seg}$. 
\subsection{Semantic-aware Instance Mask Generation}
\label{sec2}
\subsubsection{Pseudo Semantic Map}
Low-level image features, such as colors, intensity, edges, blobs, \etc, could provide useful guidance to identify the object boundaries in an image. However, these features vary significantly with illuminations, motion blurs, and noises. Thus it is error-prone to take only low-level features as supervision for BSIS when object instances are heavily blended with the background. To address this issue, we attempt to explore the intrinsic structures of objects as semantic guidance to provide more robust supervision for BSIS model training.

We construct a group of representative prototypes to model the structural information of objects, and use them to generate semantic-aware pseudo masks. Considering that a single prototype is insufficient to capture the intra-class variance, we employ multiple prototypes~\cite{ning2021multi,zhou2022rethinking} to represent the objects in a category. 
Specifically, we extract $L$ prototypes (\ie, sub-centers) from each class $c\in \{1,\cdots, C\}$, denoted by $P_c = \{p^c_1, \cdots, p^c_L\}$, to depict different characteristics of the same category. Given an input image $I\in \mathbb{R}^{h\times w \times 3}$, we first pass it through the segmentation model $\mathcal{F}_{seg}$ to obtain the feature map $Z\in \mathbb{R}^{H\times W\times D}$, and normalize it with $z_i=\frac{z_i}{||z_i||_2}$, where $z_i$ denotes the $i$-th feature vector of $Z$ with length $D$. Unlike semantic segmentation, which predicts only one mask for each input image, we predict a variable number of masks depending on the number of categories in the image. To this end, we compute the semantic probability map corresponding to the $c$-th category, denoted by $M_{\rm S}^c\in \mathbb{R}^{H\times W}$, using the following formula:
\begin{align}
	M_{{\rm S}, i}^c = \sigma ( \max\{{\frac{\langle {z}_i, {p}_{l}^c  \rangle}{\tau} \}^L_{l=1}}),
\end{align}
where {\small $\langle\cdot \rangle$} computes the cosine similarity between two $\ell_2$-normalized feature vectors. The sigmoid function {\small $\sigma(\cdot)$} converts the feature distance to the probability that the pixel belongs to the $l$-th sub-center, and $\tau$ controls the concentration level of representations. Once computed, we assign these semantic probability maps to different objects according to their class labels $Y_n$.  

\noindent \textbf{Multi-prototype update.} We update the prototypes on-the-fly with the moving average of cluster centroids computed in previous mini-batches. Specifically, given an image $X_n$ and its corresponding pseudo mask $\hat{M}$, we obtain the pixel-wise cluster assignments $Q$ of the $c$-th category by optimizing the following objective function: 
\begin{equation}
	\label{eq2}
	{\small \begin{split}
			&\underset{Q\in \mathbb{Q}}{\max}\ {\rm{Tr}}(Q^TP^{T}_cZ)+\varepsilon H(Q),\quad s.t. Q\in \mathbb{Q},\\
			{\rm with} &\quad \mathbb{Q}:=\{Q\in \mathbb{R}^{L\times N_c}_+|Q\mathbf{1}_{N_c}=r,\; Q^T\mathbf{1}_{L}=h\}.
	\end{split}}
\end{equation}
The above formula is an instance of the optimal transport problem~\cite{cuturi2013sinkhorn}, where $Q=\frac{1}{N_c}[q_1,\cdots,q_{N_c}]$ represents the transport assignment and is restricted to be a probability matrix with the constraint $\mathbb{Q}$. $N_c$ is the number of pixels belonging to the $c$-th category, $H$ denotes the entropy function with $H(Q)=-\sum_{ij}Q_{ij}\log Q_{ij}$, and $\varepsilon$ controls the smoothness of distribution. $r=\frac{1}{L}\mathbf{1}_L$ and $h=\frac{1}{N_c}\mathbf{1}_{N_c}$ are the marginal projections of $Q$ onto its rows and columns, respectively, where $\mathbf{1}_L$ and $\mathbf{1}_{N_c}$ represent the vectors of ones of dimension $L$ and $N_c$.  

By formulating the cluster assignment as an optimal transport problem, the optimization of Eq.~\ref{eq2} concerning $Q$ can be solved in linear time by the Sinkhorn-Knopp algorithm~\cite{cuturi2013sinkhorn}:
\begin{align}
	Q^* =\diag({u})\exp(\frac{P^{T}_cZ}{\varepsilon})\diag({v}),
\end{align}
where ${u} \in \mathbb{R}^L$ and ${v}\in \mathbb{R}^{N_c}$ are two renormalization vectors. Finally, we update the prototypes as the moving average of cluster centroids. Particularly, in each iteration $t$, the prototype is estimated as:
\begin{align}
	p^c_l|_t = \gamma\cdot p^c_l|_{t-1} + (1-\gamma)\cdot p^c_{n,l},
\end{align} 
where $\lambda \in [0,1]$ is the momentum coefficient. $p^c_{n,l}$ denotes the $l$-th sub-center of the $c$-th class in image $X_n$, which is computed by:
\begin{align}
	p^{c}_{n,l} = \frac{\sum_i^{N_c} z_i\cdot \mathbbm{1}(Q_{i,l}=1)}{\sum_i^{N_c}\mathbbm{1}(Q_{i,l}=1)} ,
\end{align}  
where $\mathbbm{1}$ is an indicator function, being 1 if $Q_{i,l}=1$. 

\noindent \textbf{Remarks on prototypes.} The pairwise loss used in~\cite{tian2021boxinst} explores pixel-to-pixel correlations, which provide local supervision but can not ensure the global consistency of objects with the same semantics. In contrast, the prototypes explore pixel-to-center relations, which could ensure the integrity of objects and provide more reliable supervision. Besides, since the prototypes are abstracted from massive training data, they could reveal the intrinsic properties of objects and filter out image-specific noise and outliers. In addition, we treat different categories equally and set the same number of prototypes for each category, which is potentially beneficial for identifying long-tailed objects. 
\begin{figure}
	\centering 
	\includegraphics[scale=0.14]{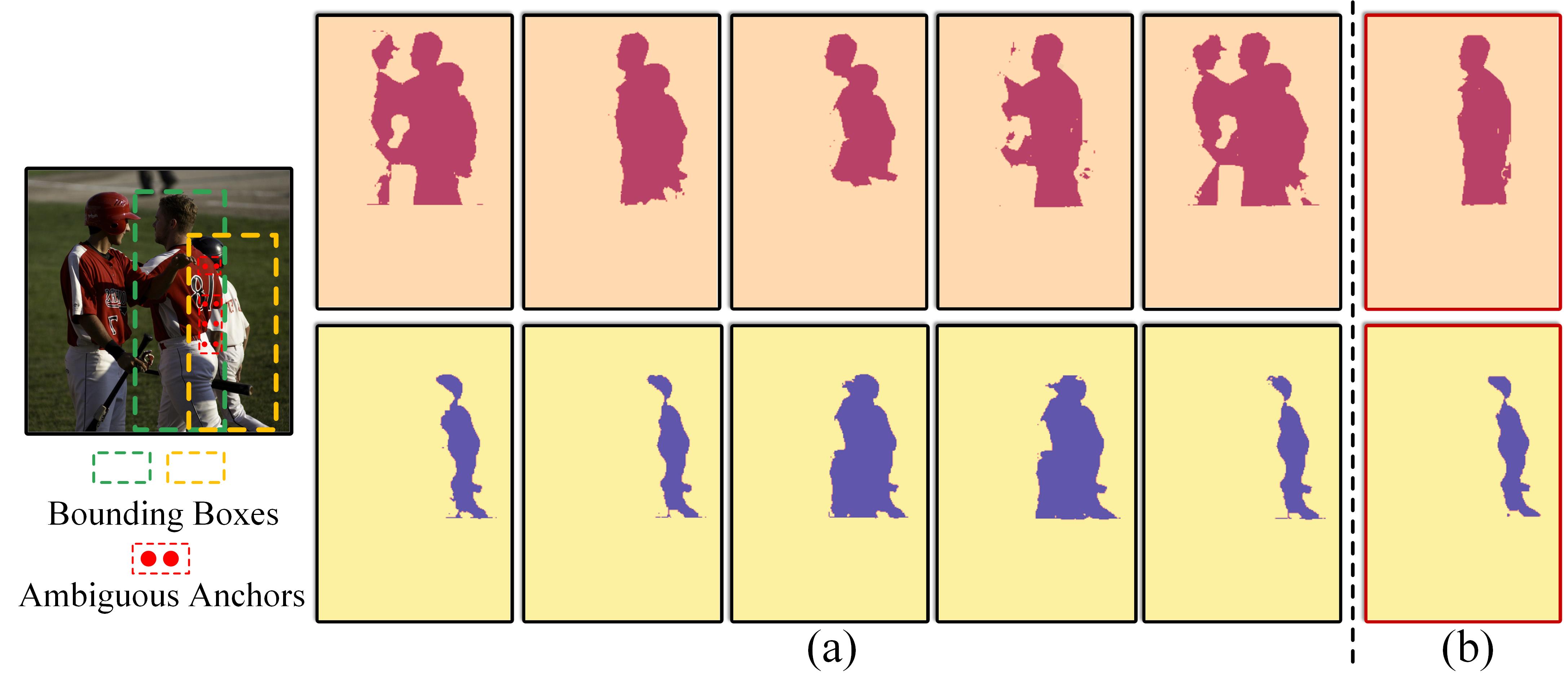}\\
	\vspace{-1.0em}
	\caption{(a) The mask quality varies much across different positive samples. (b) The instance-aware masks $M_{\rm I}$ obtained by using positive mask weighting strategy. }
	\label{fig3}
	\vspace{-1.0em}
\end{figure} 
\subsubsection{Self-Correction}
Though the pseudo semantic masks $M_{\rm S}$ could provide more reliable supervision from a global perspective, they could not distinguish different objects of the same semantics, especially when there exist overlaps or occlusions among objects. To overcome this limitation, we propose a simple yet effective self-correction module, which could upgrade the semantic-aware masks $M_{\rm S}$ to be instance-aware. 

\noindent\textbf{Positive mask weighting.}  Let us first revisit some properties of anchor-free detectors such as FCOS~\cite{tian2019fcos}. In these works, anchors denote the dense feature points, and positive samples represent the anchors located in the center/bbox region of each object. These methods assign multiple positive samples, which have high enough Intersection over Union (IoU) with ground truth (\textit{gt}) box, to each object. However, the quality of masks produced by different positive samples varies significantly, as shown in Fig.~\ref{fig3} (a). Those ambiguous anchors, \ie, anchors that are taken as positive samples for multiple \textit{gt} objects simultaneously (red dots in Fig.~\ref{fig3}), could not separate overlapping objects of the same semantics. Based on these observations, we propose a positive mask weighting strategy to integrate different masks according to their quality, resulting in a high-quality instance-aware mask $M_{\rm I}$. In specific, we define a metric of mask quality based on the IoU between predicted and \textit{gt} boxes:
\begin{align}
	w_{pos} = e^{\mu \cdot IoU},
\end{align}
where $\mu$ controls the relative gaps between different weights. Each weight $w_{pos}$ is then normalized by the sum of weights for all positive samples. As can be seen in Fig.~\ref{fig3} (b), the pseudo instance masks $M_{\rm I}$ could better separate different objects and provide more accurate supervision. 

\noindent\textbf{Pseudo mask loss.} By employing $M_{\rm I}$, the falsely activated objects or pixels in $M_{\rm S}$ could be suppressed, while the confidence of foreground objects could be enhanced. The rectification process is conducted as follows:  
\begin{align}
	\hat{M}^{k,i}_{prob} = (1-\alpha)\cdot M_{\rm S}^{k,i} + \alpha\cdot M_{\rm I}^{k,i},
	\label{eq6}
\end{align}  
where $\hat{M}_{prob}^{k,i}$ represents the $i$-th pixel of the $k$-th pseudo probability map, and $\alpha\in [0,1]$ controls the intensity of modulation. Finally, we set two thresholds $\tau_{\rm high}$ and $\tau_{\rm low}$ to select highly-confident foreground and background predictions as pseudo labels, resulting in $\hat{M}$. The pseudo-supervised mask loss is defined by:  
\begin{align}
	\mathcal{L}_{pseudo}=\frac{1}{N_{pos}}\sum_k^{N_{pos}}{\ell_{mask}(M_{pred},\hat{M}_k, W)},
\end{align}
where the mask loss $\ell_{mask}$ consists of two terms: binary cross-entropy loss $\ell_{bce}$ and dice loss~\cite{milletari2016v} $\ell_{dice}$. $\hat{M}_k$ denotes the pseudo mask of the $k$-th positive sample. $W$ is a binary weight mask that neglects ambiguous regions by using $\tau_{\rm high}$ and $\tau_{\rm low}$,~\ie, {\small $W^i=0, {\rm if}~\tau_{\rm low}<\hat{M}_{prob}^{i}< \tau_{\rm high}$}.    
\begin{figure}
	\centering 
	\includegraphics[scale=0.14]{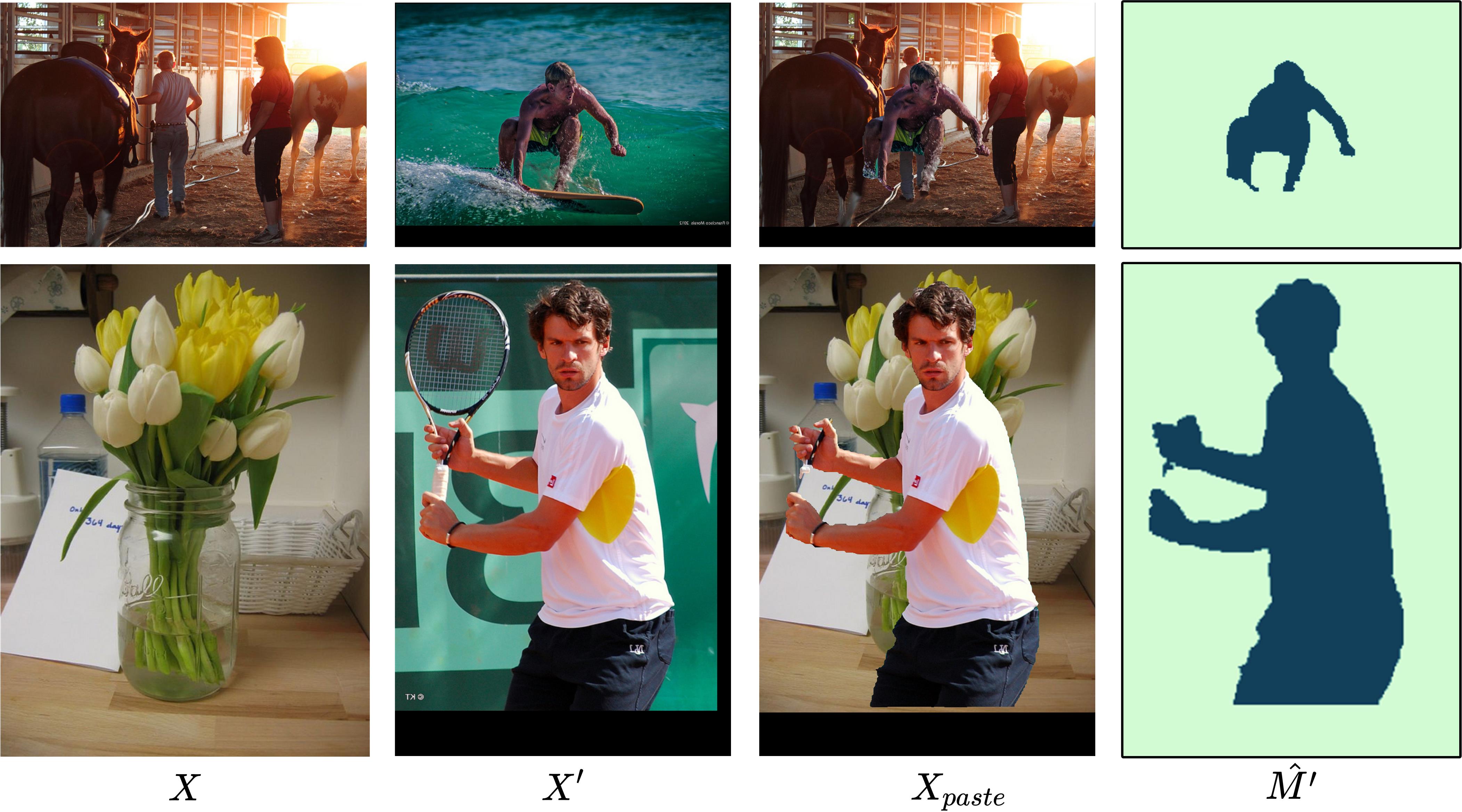}\\
	\caption{Examples of online weakly-supervised Copy-Paste. We use $\hat{M'}$ to extract instances from $X'$ and paste them onto $X$, resulting in new training data $X_{paste}$.}
	\vspace{-0.5em}
	\label{fig4}
\end{figure} 
\begin{table*}[!t]
	\centering
	\scalebox{0.85}{
		\begin{tabular}{p{0.3cm}|c|c|c|c|cc|ccc}
			\toprule\rowcolor{gray!20}
			& method            & backbone             & sche. & AP   & AP$_{50}$ & AP$_{75}$ & AP$_{S}$ & AP$_{M}$ & AP$_{L}$ \\ \hline \hline 
			\multirow{5.5}{*}{\textit{\rotatebox{90}{\small fully-supervised}}} & Mask R-CNN~\cite{he2017mask}        & ResNet-101-FPN           & $3\times$      & 37.5 & 59.3  &  40.2  &  21.1  &  39.6  & 48.3   \\
			& YOLACT-700~\cite{bolya2019yolact}        & ResNet-101-FPN           &  $4.5\times$      & 31.2 &  50.6  &  32.8  &  12.1  &  33.3  &  47.1 \\
			& PolarMask~\cite{xie2020polarmask}         & ResNet-101-FPN           &  $2\times$      & 32.1 &  53.7  &  33.1  &  14.7  &  33.8  & 45.3   \\
			& SOLOv2~\cite{wang2020solov2}            & ResNet-101-FPN           &   $3\times$    & 39.7 &  60.7  &  42.9  & 17.3   &  42.9  &  57.4  \\
			& CondInst~\cite{tian2020conditional}          & ResNet-101-FPN        & $3\times$      & 39.1 & 60.9   & 42.0   &  21.5  &  41.7  & 50.9 \\
			& Mask2Fomer$^\dag$~\cite{Cheng_2022_CVPR}          & {\small ResNet-101-MSDefomAttn}        & 50e      & 44.2  & -   & -   &  23.8  &  47.7  & 66.7 
			\vspace{0.5mm}\\ \hline
			\multirow{14}{*}{\textit{\rotatebox{90}{box-supervised}}} & BBTP$^\dag$~\cite{hsu2019weakly}              & ResNet-101-FPN   & $1\times$  &  21.1    &  45.5  &  17.2  &  11.2  &  22.0  &  29.8  \\
			& BBAM~\cite{lee2021bbam}              & ResNet-101-FPN          & $1\times$  &  25.7    &  50.0  &  23.3  &  -  & -   & -   \\
			& BoxCaseg$^\ddag$~\cite{wang2021weakly}  & ResNet-101-FPN           &  $1\times$    &   30.9   &  54.3  &  30.8  &  12.1  &  32.8  &  \textbf{46.3}  \\
			& SIM (Ours)               & ResNet-101-FPN           & $1\times$      &  \textbf{34.0}    & \textbf{56.8}   &  \textbf{35.0}  & \textbf{17.2}   &  \textbf{36.8}  &  45.5  \\\cdashline{2-10}
			& BoxLevelSet~\cite{li2022box}       & ResNet-101-FPN           & $3\times$    &  33.4   &  56.8  &  34.1  &  15.2  &  36.8  &  46.8  \\
			& BoxInst~\cite{tian2021boxinst}           & ResNet-101-FPN           & $3\times$      &   33.2   &  56.5  &  33.6  &  16.2  &  35.3  &  45.1  \\
			& SIM (Ours)              & ResNet-101-FPN           & $3\times$     &   \textbf{35.3}   &  \textbf{58.9}  &  \textbf{36.4}  &  \textbf{18.4}  &  \textbf{38.0}  &  \textbf{47.5}  \\ \cdashline{2-10}
			& BoxLevelSet~\cite{li2022box}       & {\small ResNet-DCN-101-BiFPN}          & $3\times$      &   35.4   &  59.1  &  36.7  &  16.8  &  38.5  &  {51.3}  \\
			& BoxInst~\cite{tian2021boxinst}           & {\small ResNet-DCN-101-BiFPN} & $3\times$      &   35.0   &  59.3  &  35.6  &  17.1  &  37.2  &  48.9  \\
			& SIM (Ours)              & {\small ResNet-DCN-101-BiFPN} & $3\times$      &   \textbf{37.4}   &  \textbf{61.8}  &  \textbf{38.6}  &  \textbf{18.6}  &  \textbf{40.2}  &  \textbf{51.6}  \\ \cdashline{2-10}
			& BoxInst~\cite{tian2021boxinst}           & Swin-B-FPN               & $3\times$      &  37.9    &  63.2  &  39.0  &  20.0  &  41.2  &  53.1  \\
			& SIM (Ours)              & Swin-B-FPN               &  $3\times$     &   \textbf{40.2}   &  \textbf{66.9}  &  \textbf{41.3}  &  \textbf{21.1}  &  \textbf{43.5}  &  \textbf{56.0}  \\ 
			\cdashline{2-10}
			& BoxInst$^\dag$~\cite{tian2021boxinst}          & {\small Mask2Former-ResNet-101}              & 50e      &  35.7    &  59.8  & 36.4 &  16.6  & 38.5  & 55.4  \\
			& SIM$^\dag$ (Ours)              & {\small Mask2Former-ResNet-101}              &  50e   &  \textbf{37.4}  &  \textbf{62.2}  & \textbf{ 38.7}  & \textbf{ 17.6}  & \textbf{ 41.3 } &  \textbf{56.6 } \\ 
			\bottomrule
	\end{tabular}}
	\caption{Comparisons between SIM and  state-of-the-art methods on the COCO \texttt{test-dev} split. Symbol ``\dag'' means that the results are evaluated on the COCO \texttt{val} split, and ``\ddag'' denotes that BoxCaseg is trained with both box and salient object supervisions.}
	\label{tab1}
\end{table*} 
\subsection{Online Weakly-Supervised Copy-Paste} 
\label{sec3}
Object-aware Copy-Paste is a simple yet effective way to improve the data efficiency. However, Copy-Paste has rarely been explored for weakly-supervised instance segmentation. It is natural to employ pseudo masks as the guidance to cut object instances from an image $X$. To achieve online Copy-Paste, we set up a first-in-first-out memory bank $\mathcal{M}$ to store training samples and their corresponding pseudo masks from preceding mini-batches, which ensures that the pseudo masks in $\mathcal{M}$ could be updated on-the-fly.

For each training iteration, we randomly sample an image $\{X', Y', B', \hat{M}',S'\}$ from $\mathcal{M}$ and extract a subset of instances from $X'$ based on importance sampling, where $S'$ measures the importance of instances (please refer to \textbf{supplemental materials} for more details), so that instances with higher-quality masks are more likely to be selected. We paste the extracted objects onto input image $\{X, Y, B\}$, and adjust the annotations accordingly, \ie, we remove fully occluded objects and update the masks and bounding boxes of partially occluded objects. Finally, we compute the mask loss only on the pasted instances:
\begin{align}
	{\small 	\mathcal{L}_{paste} = \sum_k \mathbbm{1}_{paste}[\ell_{mask}(M^k_{pred},\hat{M}'^k)],}
\end{align} where $\mathbbm{1}_{paste}$ is the indicator function, being $1$ if the $k$-th instance is copied from $X'$.

\subsection{Objective Function}
As shown in Fig.~\ref{fig2}, we employ a momentum encoder to stabilize the pseudo mask generation process. The parameters of the segmentation model are updated by optimizing the following loss function $\mathcal{L}_{seg}$:
\begin{align}
	\mathcal{L}_{seg} = \mathcal{L}_{lowlevel} + \lambda_1\mathcal{L}_{pseudo} + \lambda_2\mathcal{L}_{paste},
\end{align} 
where $\lambda_1$ and $\lambda_2$ are two trade-off parameters. $\mathcal{L}_{lowlevel}$ denotes low-level pairwise supervision defined in BoxInst~\cite{tian2021boxinst}. $\mathcal{L}_{lowlevel}$ and $\mathcal{L}_{pseudo}$ provide complementary supervision from local and global perspectives, respectively, and work together to bridge the performance gap between box-supervised and fully-supervised settings.

\section{Experiments}
We conduct experiments on COCO~\cite{lin2014microsoft} and PASCAL VOC~\cite{everingham2010pascal} datasets. The model is trained on {\small \texttt{train2017}}, which contains about 115k images from 80 categories with only box annotations. We use {\small \texttt{val2017}} (5k images) for ablation study and {\small \texttt{test-dev2017}} (20k images) for comparisons with other methods. 
\subsection{Implementation Details}
We adopt CondInst~\cite{tian2020conditional} and Mask2Former~\cite{Cheng_2022_CVPR} as our baseline. For CondInst, the backbone with FPN is pre-trained on ImageNet. The training and testing details follow CondInst\footnote{https://github.com/aim-uofa/AdelaiDet} implemented with {\small \texttt{Detectron2}}~\cite{wu2019detectron2} unless specified. The model is warmed-up for 10k iterations with the projection loss and pairwise loss proposed in~\cite{tian2021boxinst}, and then trained for 80k iterations by adding our pseudo supervision loss and Copy-Paste loss with batch size 16 on 8 TITAN RTX GPUs. 
When ResNet is used as the backbone, our model is trained with SGDM optimizer. The initial learning rate is set to 0.01, and reduced by a factor of 10 at steps 60k and 80k, respectively. When SwinT~\cite{liu2021swin} is used as the backbone, we adopt the AdamW~\cite{loshchilov2018decoupled} optimizer and set the initial learning rate to 0.0001. For Mask2Former, we follow its baseline settings\footnote{https://github.com/facebookresearch/Mask2Former} and replace the original pixel-wise mask loss with our designed loss terms. The length of the memory bank is set to 100, and we extract a quarter of the instances from each image with $1\sim3$ instances per image. The momentum used to update networks and prototypes is set to 0.9999 and 0.999, respectively. The modulation intensity $\alpha$ is empirically set to 0.5. Besides, $\lambda_1$, $\lambda_2$, $\mu$, and $\tau$ are empirically set to $0.5$, $1$, $5$, and $0.1$, respectively.   
\begin{table}[!t]
	\centering
	\scalebox{0.85}{
		\begin{tabular}{c|c|c|cc}
			\toprule\rowcolor{gray!20}
			methods     & backbone & AP & AP$_{50}$ & AP$_{75}$ \\ \hline\hline
			GrabCut$^*$~\cite{rother2004grabcut}     &  ResNet-101        &  19.0  & 38.8 & 17.0 \\
			SDI~\cite{khoreva2017simple}         &  VGG-16        &  -  & 44.8 & 16.3 \\
			BBTP~\cite{hsu2019weakly}        &   ResNet-101       &  23.1  & 54.1 & 17.1 \\
			BBTP w/ CRF~\cite{hsu2019weakly} &    ResNet-101      &  27.5  & 59.1 & 21.9 \\
			BBAM~\cite{lee2021bbam}        &  ResNet-101        &  -  & 63.7 & 31.8 \\
			BoxInst~\cite{tian2021boxinst}     &  ResNet-50        &  34.3  & 59.1 & 34.2 \\
			BoxInst~\cite{tian2021boxinst}     &  ResNet-101        &  36.5  & 61.4 & 37.0 \\
			DiscoBox~\cite{lan2021discobox}    &  ResNet-50        &  -  & 59.8 & 35.5 \\
			DiscoBox~\cite{lan2021discobox}    &  ResNet-101        &  -  & 62.2 & 37.5 \\
			BoxLevelSet~\cite{li2022box} &  ResNet-50        &  36.3  & 64.2 & 35.9 \\
			BoxLevelSet~\cite{li2022box} &  ResNet-101        &  38.3  & 66.3 & {\bf 38.7} \\ \hline
			SIM (Ours)   &   ResNet-50       &  36.7  & 65.5 & 35.6 \\
			SIM (Ours)   &   ResNet-101      &  {\bf{38.6}}  & {\bf 67.1} & 38.3 \\ \bottomrule
	\end{tabular}}
	\caption{Performance comparison on Pascal VOC \texttt{val2012} split. Symbol ``$*$" denotes that the results are copied from BoxInst.} 
	\label{tab2}
	\vspace{-1.0em}
\end{table}
\begin{figure*}
	\centering 
	\includegraphics[scale=0.21]{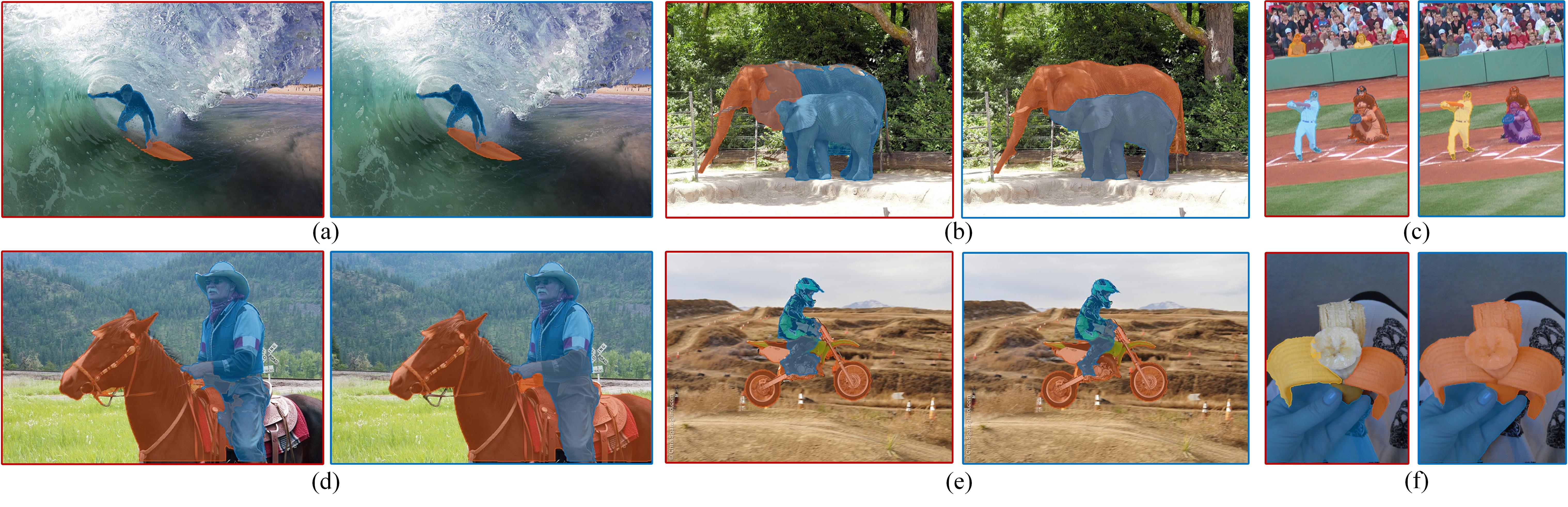}\\
	\vspace{-1em}
	\caption{Qualitative results of BoxInst (in the {\color{red}{red}} box) and our method (in the {\color{blue}{blue}} box) on COCO \texttt{val2017}.}
	\label{fig5}
	\vspace{-0.5em}
\end{figure*} 

\subsection{Comparisons with State-of-the-Arts}
We compare the proposed SIM with state-of-the-art BSIS methods on the COCO~\cite{lin2014microsoft} dataset. The fully supervised methods are also compared as a reference. As shown in Tab.~\ref{tab1}, SIM outperforms BoxInst~\cite{tian2021boxinst} and BoxLevelSet~\cite{li2022box} by 1.6\% and 1.4\% AP with the ResNet-101-FPN backbone and $3\times$ training schedule. This is because we employ dataset-level prototypes to exploit the semantic information of objects while filtering out trivial image-specific noises. It can also be seen that our SIM method produces impressive results on small objects, largely outperforming BoxInst and BoxLevelSet by 2.8\% AP and 1.8\% AP with ResNet-101-FPN backbone, respectively. This can be attributed to the proposed Copy-Paste operation, which creates many challenging training data of small hard objects. BoxInst lags behind on large objects due to the lack of semantic guidance. BoxLevelSet has lower performance on small objects because small objects lack rich features for level set evolution. By using stronger backbones with BiFPN~\cite{tan2020efficientdet} and DCN~\cite{zhu2019deformable}, the performance can be further boosted to 37.4\% AP. By taking Swin transformer~\cite{liu2021swin} as backbone~\cite{liu2021swin}, the proposed SIM could surpass BoxInst by 1.7\% AP, attaining 39.6\% AP. In addition, we also validate the effectiveness of our method on the query-based baseline,~\ie, Mask2Former~\cite{Cheng_2022_CVPR}. Our method achieves consistent improvement and outperforms BoxInst by 1.7\% AP. 

Tab.~\ref{tab2} reports the segmentation results on the Pascal VOC~\cite{everingham2010pascal} dataset. Our method outperforms BoxInst\cite{tian2021boxinst} by 2.4\% and 2.1\% AP with ResNet-50 and ResNet-101 backbones, respectively. BoxLevelSet~\cite{li2022box} achieves comparable performance since the level set model could evolve the precise contour of objects, which is beneficial for large objects.
\subsection{Qualitative Results}
Fig.~\ref{fig5} shows the qualitative segmentation results of our method and BoxInst on COCO~\textit{val} split. We have the following observations. First, according to (a), by leveraging the semantic-level supervision, our method is able to segment foreground instances that heavily tangle with background or other objects with similar appearances, because the prototypes explore global structural information of objects and could reduce the noise brought by using only local pair-wise affinity supervision. Second, as can be seen in (b) (c), our method could better separate overlapping instances of the same semantics since we introduce the self-correction module to reduce the falsely activated instances while enhancing the correct ones. Third, as shown in (d) and (f), benefiting from the abstraction of prototypes from massive training data, our method is able to perceive the whole entity of object instances and produce better segmentation results. 

\begin{table}[!t]
	\centering 
	\scalebox{0.78}{
		\begin{tabular}{cc|c|cc|ccc}
			\toprule\rowcolor{gray!20}
			\multicolumn{1}{c|}{$\mathcal{L}_{pseudo}$} & $\mathcal{L}_{paste}$ & AP & AP$_{50}$ & AP$_{75}$ & AP$_{S}$ & AP$_{M}$ & AP$_{L}$ \\ \hline\hline
			\multicolumn{2}{c|}{baseline}         &  30.7  & 52.2 & 31.1 & 13.8 & 33.1 & 45.7 \\ \hline
			\multicolumn{1}{c|}{\checkmark}   &    &  31.9  & \textbf{54.0}   & 32.6 & 14.7 & \textbf{34.7} & 47.4   \\
			\multicolumn{1}{c|}{\checkmark}       & \checkmark &  \textbf{32.2}  & \textbf{54.0} & \textbf{33.0} & \textbf{15.8} & 34.5 & \textbf{48.3} \\ \bottomrule
	\end{tabular}}
	\vspace{-0.7em}
	\caption{The Mask AP on COCO \texttt{val2017} split by applying different loss terms. }
	\vspace{-0.5em}
	\label{tab3}
\end{table}
\begin{table}[!t]
	\centering
	\scalebox{0.75}{
		\begin{tabular}{p{8mm}|p{10mm}|p{10mm}p{10mm}|p{10mm}p{10mm}p{10mm}}
			\toprule\rowcolor{gray!20}
			$\alpha$ & AP & AP$_{50}$ & AP$_{75}$ & AP$_{S}$ & AP$_{M}$ & AP$_{L}$ \\ \hline\hline
			$0$   &  30.5  &  52.7  & 30.6 & 14.1 & 33.3 &  44.8\\
			$0.3$ &  31.3  &  53.1  & 31.9 & 15.1 & 34.0 &  46.1 \\
			$0.5$ &  \textbf{32.2}  &  \textbf{54.0}  & \textbf{33.0} & \textbf{15.8} & \textbf{34.5} & \textbf{48.3} \\
			$0.7$ &  32.0  & \textbf{54.0} & 32.4 & 15.7 & 34.4 & 47.9 \\
			$1$   &  31.4  &  53.0  & 32.3 & 15.0 & 34.2 & 47.0 \\
			\bottomrule
	\end{tabular}}
	\vspace{-0.7em}
	\caption{Effect of modulation intensity $\alpha$.}
	\label{tab4}
	\vspace{-0.5em}
\end{table}
\begin{table}[]
	\centering
	\scalebox{0.75}{
		\begin{tabular}{c|p{0.8cm}|p{0.8cm}p{0.8cm}|p{0.9cm}p{0.9cm}p{0.9cm}}
			\toprule\rowcolor{gray!20} 
			\# prototypes & AP & AP$_{50}$ & AP$_{75}$ & AP$_{S}$ & AP$_{M}$ & AP$_{L}$ \\ \hline\hline
			$L=1$           &  31.6  &  53.1  &  32.4  &  15.3  &  34.0  &  46.9  \\
			$L=5$          &  32.0  &  53.6  &  32.9  &  15.7  &  \textbf{34.5}  &  47.4  \\
			$L=10$          &  \textbf{32.2}  &  \textbf{54.0}  &  33.0  &  15.8  &  \textbf{34.5}  &  48.3  \\
			$L=50$          &  \textbf{32.2}  &  53.9  &  \textbf{33.2}  &  \textbf{15.9}  &  34.3  & \textbf{48.7}  \\ \bottomrule
	\end{tabular}}
	\vspace{-0.7em}
	\caption{Effect of the number of prototypes $L$ per category.}
	\label{tab5}
	\vspace{-1.0em}
\end{table}

\subsection{Ablation Study}
We conduct ablation studies on the COCO dataset, with ResNet-50-FPN backbone and $1\times$ training schedule, to investigate the role of each component in our framework. The Mask AP on COCO \texttt{val} split is reported. 

\begin{figure*}
	\centering 
	\includegraphics[scale=0.16]{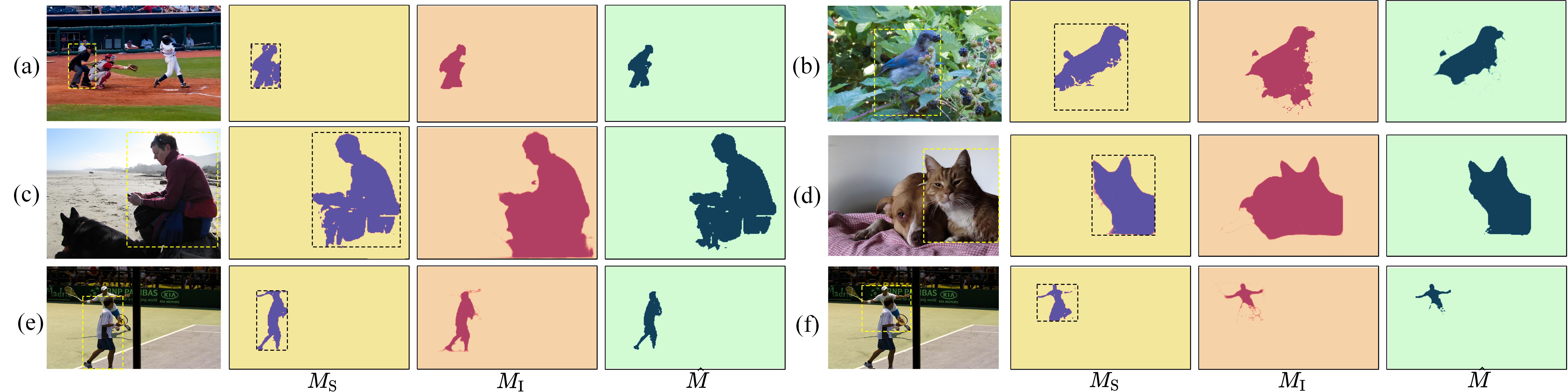}\\
	\vspace{-0.7em}
	\caption{Visualizations of pseudo semantic masks $M_{\rm S}$, pseudo instance masks $M_{\rm I}$, and final pseudo masks $\hat{M}$.}
	\vspace{-0.7em}
	\label{mask}
\end{figure*} 
\begin{figure}
	\centering 
	\includegraphics[scale=0.41]{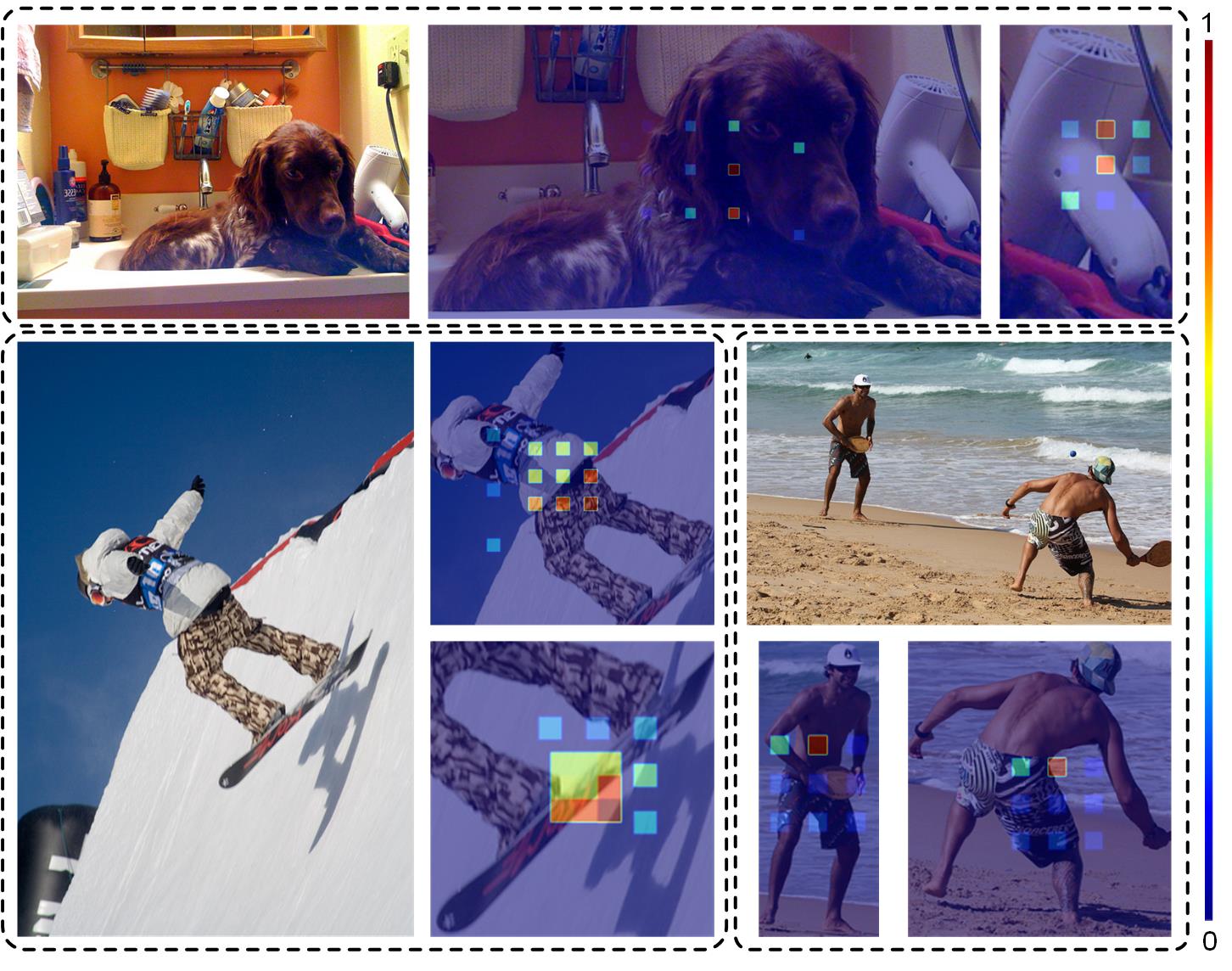}\\
	\vspace{-0.5em}
	\caption{Visualizations of weights for different positive samples.}
	\vspace{-1.3em}
	\label{weight}
\end{figure}

\noindent\textbf{Contribution of the two losses.}
Tab.~\ref{tab3} shows the contributions of the two loss terms, \ie, $\mathcal{L}_{pseudo}$ and $\mathcal{L}_{paste}$. The baseline is BoxInst~\cite{tian2021boxinst}, which resorts to the color similarity between proximal pixels as supervision. The proposed pseudo mask loss improves the performance by 1.2\% AP, especially on large objects (1.7\% AP). This demonstrates that by introducing the semantic-aware instance mask generation paradigm, our method could distinguish not only proximal pixels with similar colors but also overlapping objects of the same semantics. Besides, the online weakly-supervised Copy-Paste loss brings a further improvement of 0.3\% AP, while the performance on small objects $\textrm{AP}_{S}$ is largely improved by 1.1\% AP. 

\noindent\textbf{Effect of the modulation intensity.} We use the instance-aware pseudo probability map $M_{\rm I}$ to modulate the semantic-aware map $M_{\rm S}$ by using a parameter $\alpha$. Tab.~\ref{tab4} shows the results by setting $\alpha$ to different values. One can draw a conclusion that the integration of semantic mask $M_{\rm S}$ and instance mask $M_{\rm I}$ leads to better results than any of them. Specifically, $M_{\rm S}$ facilitates more holistic object pattern understanding by exploring the semantic information from the entire dataset, while $M_{\rm I}$ rectifies the falsely activated objects in $M_{\rm S}$ and improves the reliability of pseudo masks. Removing the self-correction module, \ie, setting $\alpha$ to 0, will lead to an obvious performance drop by 1.7\% AP. This is not surprising because the semantic mask contains certain noisy supervision caused by the falsely activated objects and pixels. On the other hand, training without semantic masks, \ie, setting $\alpha$ to 1, will decrease the segmentation performance by 0.8\% AP. 


\noindent\textbf{Semantic masks \textit{vs.} instance masks.} We explore the complementarity of $M_{\rm S}$ and $M_{\rm I}$ by visualizing them in Fig.~\ref{mask}. As can be seen in (b-d), $M_{\rm S}$ could distinguish object instances that have similar appearances but different semantics from the background and other objects. In comparison, from (a) (e-f), we see that $M_{\rm I}$ is good at distinguishing different instances of the same semantics, which are employed to suppress the falsely activated instances in $M_{\rm S}$. They work together to produce more reliable supervision for training.   

\noindent\textbf{Visualizations of weights.} To further understand the effect of the positive mask weighting strategy, we show the visualizations of weights assigned to different positive samples in Fig.~\ref{weight}. It can be seen that large weights are mainly located in the central regions of foreground objects and far from other objects, while the small ones are located in ambiguous regions, such as overlapping regions between different objects and junctions between foreground and background. This weighting strategy facilitates generating higher-quality supervision and reducing the falsely activated instances. More analyses about this weighting strategy are provided in \textbf{supplemental files}.

\noindent\textbf{Effect of the number of prototypes per category.} We set multiple prototypes per category to better model intra-class variation. Tab.~\ref{tab5} reports the segmentation results w.r.t. different number of prototypes per category. The baseline attains 31.6\% AP by representing each category with one prototype. There is a clear performance improvement (about 0.4\% AP) by increasing the number of prototypes to $5$; however, the performance reaches saturation when $L$ is more than $10$. We set $L$ as $10$ to trade off accuracy and cost.

\section{Conclusion}
We proposed a novel SIM method for box-supervised instance segmentation. To alleviate the limitations of pair-wise affinity supervision from low-level image features, we explored high-level image semantic contexts by extracting a group of representative prototypes from the dataset and using them to segment foreground objects from background. To rectify the possible false positive instances in semantic masks, we introduced extra supervision by integrating mask predictions of different positive samples in a weighted manner. Furthermore, we devised an online weakly-supervised Copy-Paste method to create challenging training data by equipping a continuously updated memory bank to store historical images with pseudo masks. Both the qualitative and quantitative experiments demonstrated the superior performance of our SIM method over state-of-the-arts. 

\clearpage
\section{Supplemental Materials}
Our supplemental file contain the following materials:
\begin{itemize}
	\item[$\bullet$] Details about the online weakly-supervised Copy-Paste operation, and some visualization results (\cf Sec3.3-Online Weakly-Supervised Copy-Paste in the main paper);
	\item[$\bullet$] Implementation details on Mask2Former~\cite{cheng2022masked} (\cf Sec4.1-Implementation Details in the main paper);
	\item[$\bullet$] Analysis on positive weighting strategy and more visualizations (\cf Sec3.2.2-Positive Mask Weighting and Sec4.4-Visualizations of Weights in the main paper); 
	\item[$\bullet$] More qualitative results (\cf Sec4.3-Qualitative Results in the main paper);
	\item[$\bullet$] More parameter analyses (\cf Sec4.4-Ablation Study in the main paper).
\end{itemize}

\subsection{Online Weakly-Supervised Copy-Paste}
Copy-paste is a simple yet effective way to improve the data efficiency of instance segmentation models. By pasting objects of various categories and scales to different images, copy-paste could achieve solid performance improvements on strong baseline models~\cite{dvornik2018modeling, dwibedi2017cut,fang2019instaboost,ghiasi2021simple}. In addition, Ghiasi~\etal~\cite{ghiasi2021simple} demonstrated the efficacy of copy-paste under the semi-supervised learning setting. However, Copy-Paste has rarely been explored for weakly-supervised instance segmentation. In this work, we use copy-paste to create new training data for better handling the object occlusions and rare object categories. Following the work~\cite{ghiasi2021simple}, we adopt a simple strategy of randomly picking objects and pasting them on the target image, which could provide a considerable boost on top of baselines. 
\begin{figure*}
	\centering 
	\includegraphics[scale=0.21]{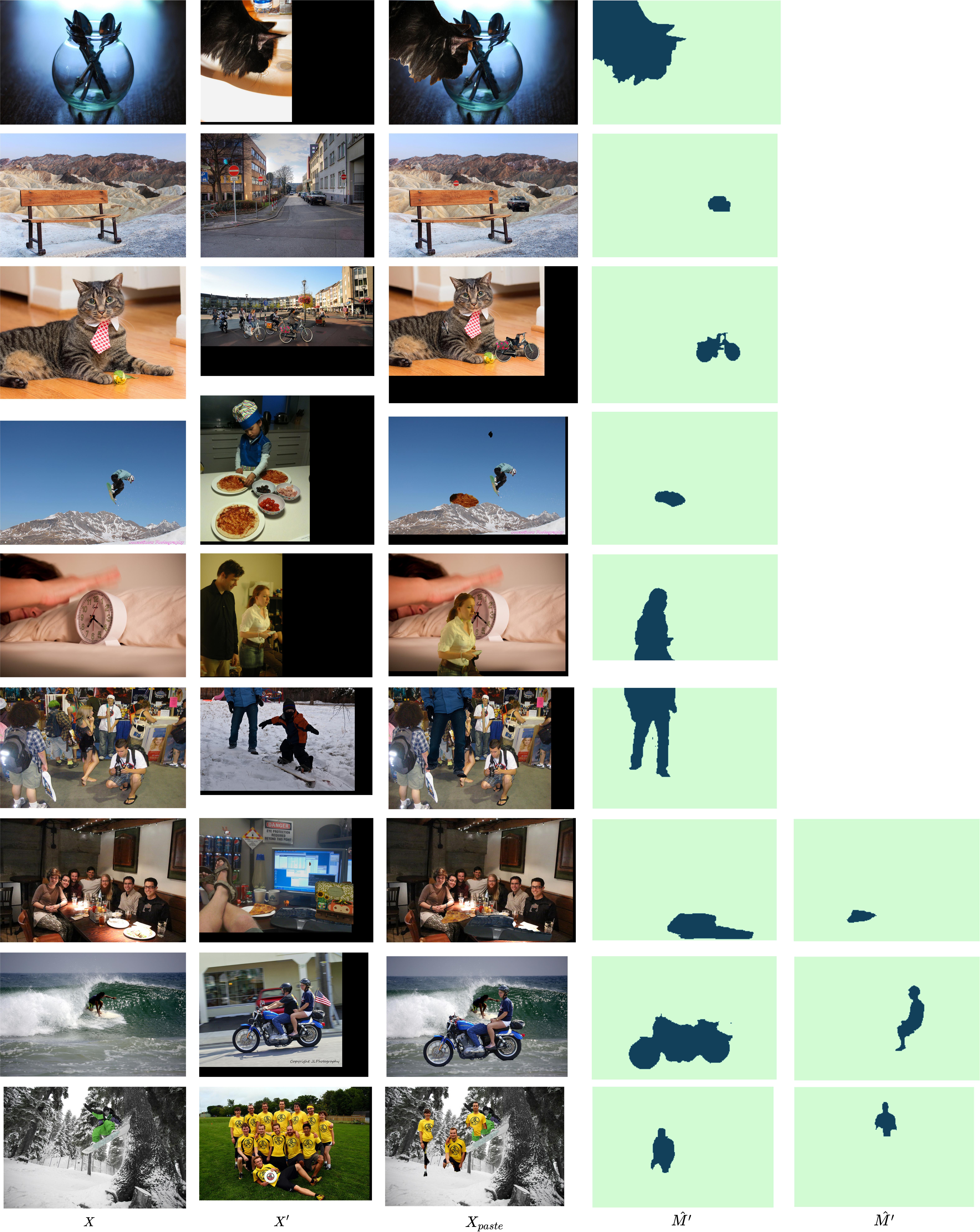}\\
	\caption{Examples of online weakly-supervised Copy-Paste. We use $\hat{M}'$ to extract instances from $X'$ and paste them onto $X$, resulting in new training data $X_{paste}$.}
	\label{paste}\vspace{-1.0em}
\end{figure*}

\begin{figure*}
	\centering 
	\includegraphics[scale=1.0]{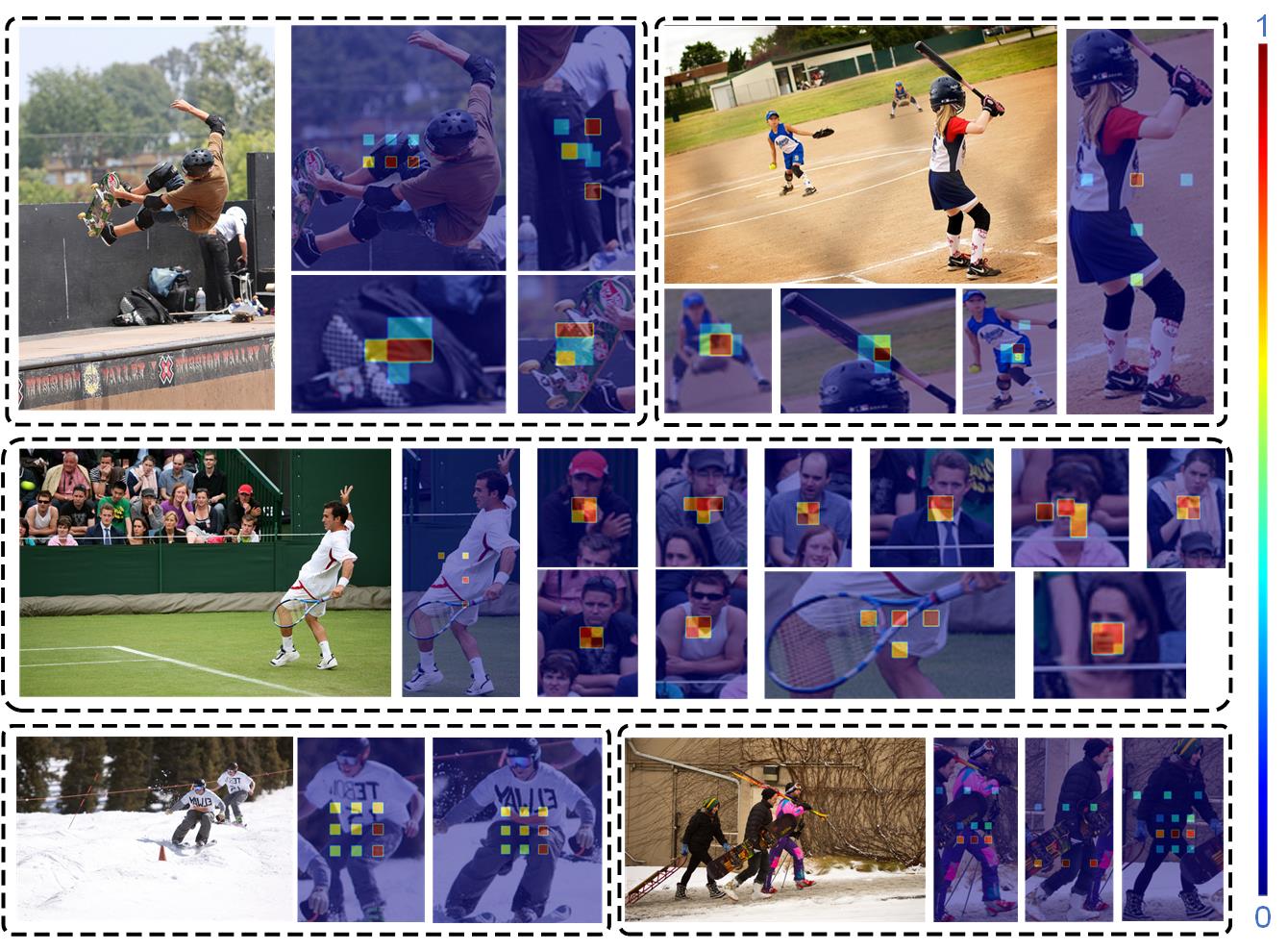}\\
	\vspace{-1.0em}
	\caption{Visualizations of weights for different positive samples.}
	\label{weight_2}
\end{figure*}


{\bf Importance Sampling.} The quality of pseudo mask varies significantly across different instances within an image. A poor mask may produce an unconvincing paste, while a good mask could result in a convincing one. Therefore, we adopt an importance sampling strategy to select instances with high-quality masks. Specifically, we calculate the averaged mask score $S$ that reflects the importance of each instance by: 
\begin{align}
	S_k = \frac{ {\sum_{i}^{N_k}M_{prob}^{k,i}\cdot \mathbbm{1}(\hat{M}^{k,i}=1)} }{\sum_{i}^{N_k}\mathbbm{1}(\hat{M}^{k,i}=1) } ,
\end{align}
where $N_k$ denotes the number of pixels from the $k$-th instance map of $M_{prob}$, and $S_k$ is the score of the $k$-th instance. We also store $S$ into the memory bank $\mathcal{M}$.

For each training iteration, we randomly sample an image $\{X',Y', B', \hat{M}',S'\}$ from $\mathcal{M}$ and randomly extract a subset of instances according to their mask scores $S'$, so that the instances with higher-quality masks are more likely to be selected. 

{\bf Examples of online weakly-supervised Copy-Paste} are shown in Fig.~\ref{paste}, we cut objects which have significant contrast with the background, and randomly paste them on the target image. In this way, more challenging data with various occlusion patterns can be created, which could effectively improve the model ability to handle the occlusions between objects. In addition, since small objects are the majority (more than 40\%) in the COCO dataset, more small objects will be created so that the Copy-Paste strategy will benefit small objects much (please refer to the Tab.~\textcolor{red}{1} in our main paper).   

\begin{figure*}
	\centering 
	\includegraphics[scale=0.70]{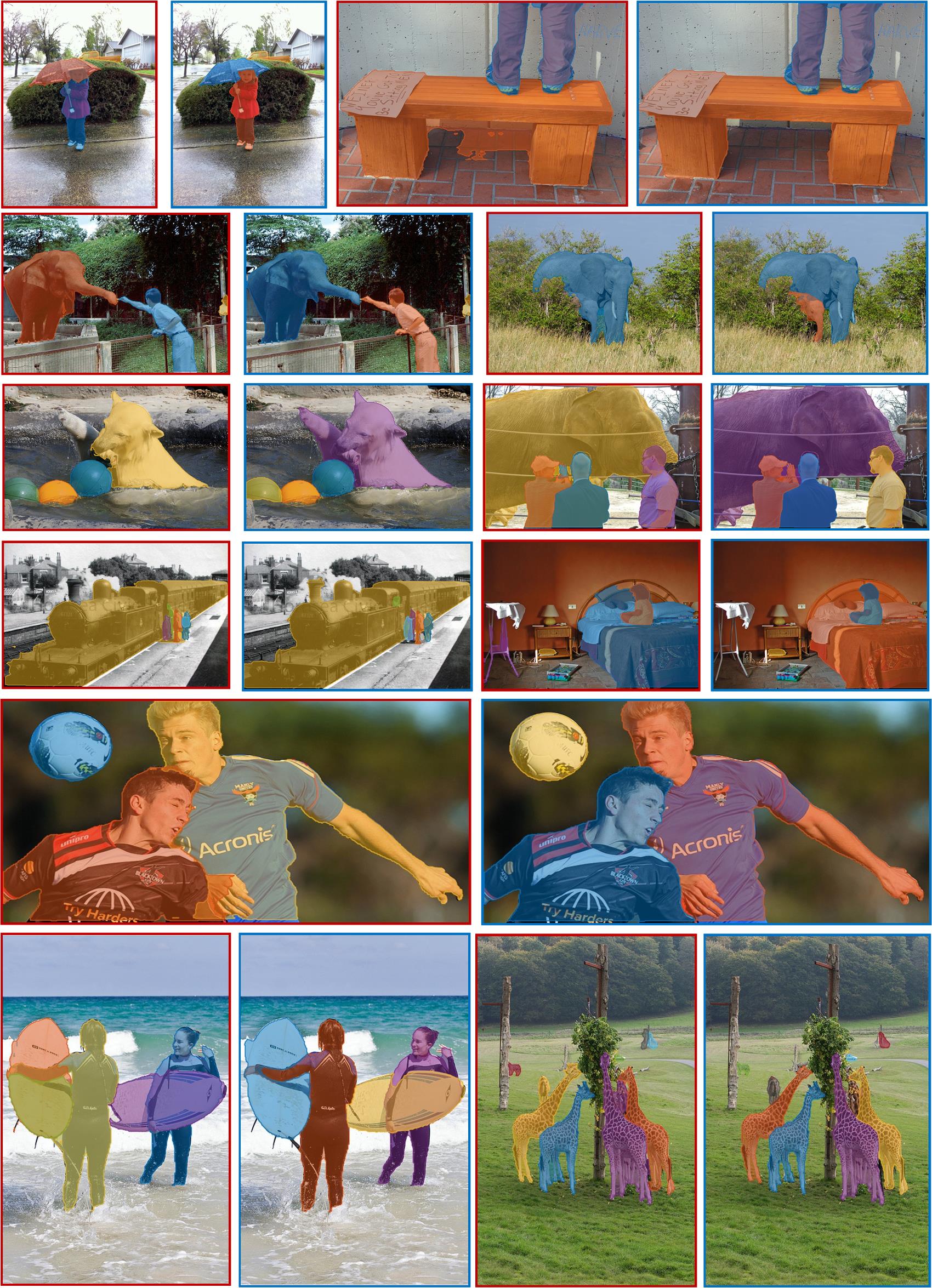}\\
	\caption{Qualitative results of BoxInst~\cite{tian2021boxinst} (in the \textcolor{red}{red} box) and our method (in the \textcolor{blue}{blue} box) on COCO \texttt{val2017}.}
	\label{vis}\vspace{-1.0em}
\end{figure*}

\begin{figure*}
	\centering 
	\includegraphics[scale=0.33]{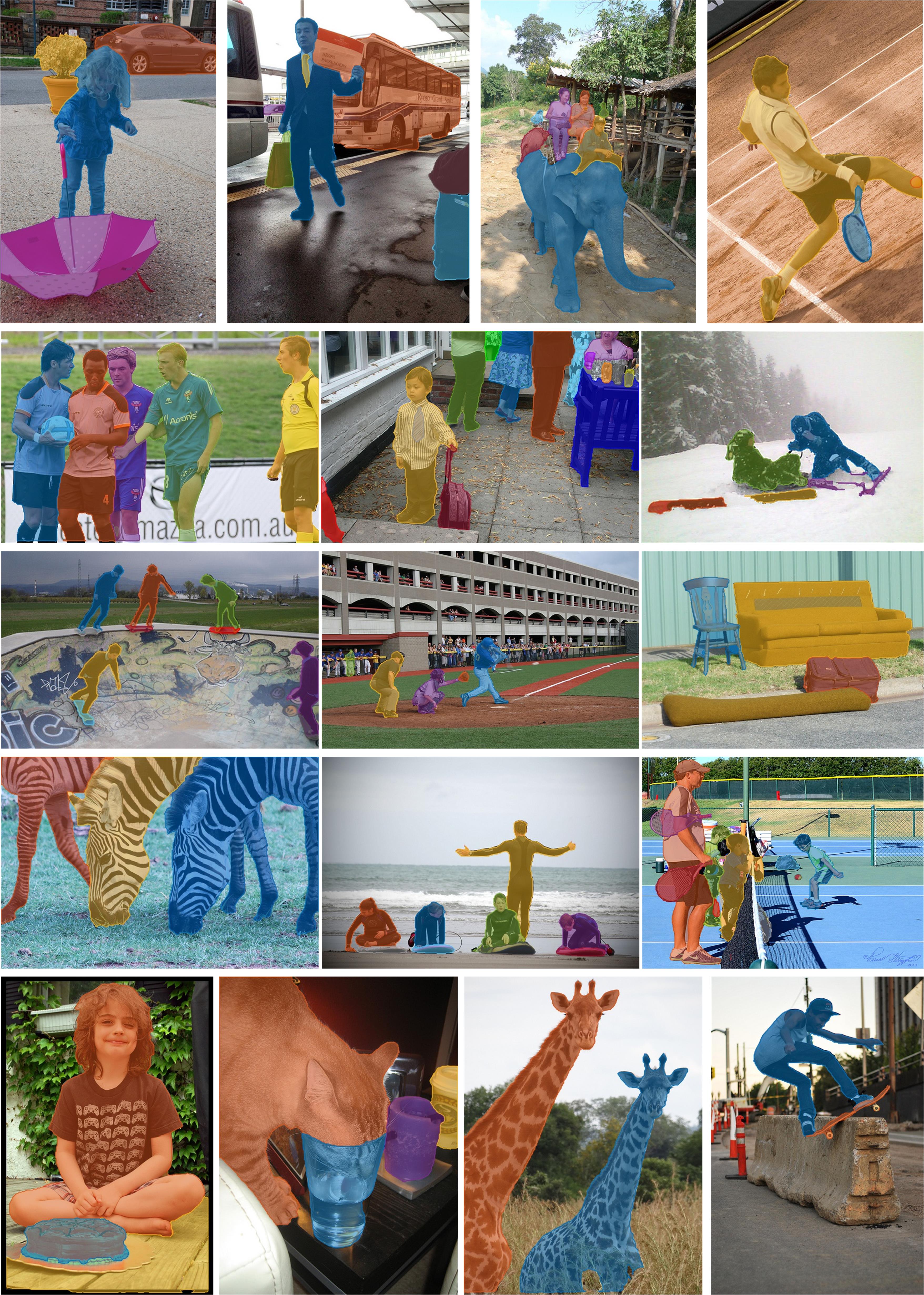}\\
	\caption{Qualitative results of our method on COCO \texttt{val2017}.}
	\label{vis2}\vspace{-1.0em}
\end{figure*}

\subsection{Implementation Details}
We adopt CondInst~\cite{tian2020conditional} and Mask2Former~\cite{cheng2022masked} as our baselines. For Mask2former, we use Detectron2 and follow the commonly used settings for each dataset. More specifically, we use AdamW~\cite{loshchilov2017decoupled} and the \textit{poly}~\cite{chen2017deeplab} learning rate schedule with an initial learning rate of $10^{-4}$. A learning rate multiplier of $0.1$ is applied to CNN backbones. The more advanced multi-scale deformable attention transformer~\cite{zhu2020deformable} is employed as the default pixel decoder.

\subsection{Analysis on Positive Weights}
In Sec.~3.2.2 of our main paper, we use the instance map to rectify the semantic map so that those falsely activated objects and regions will be suppressed while the correct ones are enhanced. Since the quality of masks produced by different positive samples varies significantly, we propose a positive mask weighting strategy to integrate different masks according to their quality. We visualize the weights of different positive samples in Fig.~\ref{weight_2}. One can observe that large weights tend to locate in the central regions of foreground objects, and the smaller weights tend to locate in ambiguous regions. 

{\bf Effect of $\mu$.} The parameter $\mu$ in Eq.~6 of our manuscript controls the relative gaps between different weights. We investigate its effect in Tab.~\ref{mu}. Out method attains the best performance when $\mu$ is set to $5.0$. When we equally treat different positive samples (\ie, set $\mu$ to 0), the performance decreases by 0.8\% AP. When we set $\mu$ to a larger value, \eg, $\mu=20$, the final result drops by 0.9\% AP. This is because we rely on some certain positive samples while ignoring other useful ones. Actually, different positive samples provide complementary information and they should be fully integrated for producing better results. 
\begin{table}[!h]
	\centering
	\scalebox{0.8}{
		\begin{tabular}{c|c|cc|ccc}
			\toprule\rowcolor{gray!20}
			$\mu$ & AP & AP$_{50}$ & AP$_{75}$ & AP$_{S}$ & AP$_{M}$ & AP$_{L}$ \\ \hline\hline
			0   &  31.4  &  52.9  & 31.6 & 14.3 & 33.7 &  46.9\\
			1  &  31.6  &  53.4  & 32.0 & 14.5 & 34.1 &  47.2 \\
			5 &  {32.2}  &  {54.0}  & {33.0} & {15.8} & 34.5 & {48.3} \\
			10 &  31.8  & 53.9 & 32.3 & 15.0 & {34.8} & 46.8 \\
			20   &  31.3  &  53.1  & 31.9 & 15.1 & 33.9 & 46.1 \\
			\bottomrule
	\end{tabular}}
	\vspace{-0.7em}
	\caption{Effect of parameter $\mu$.}
	\label{mu}
	\vspace{-0.5em}
\end{table}
\begin{table}[]
	\centering
	\scalebox{0.8}{
		\begin{tabular}{cc|c|cc|ccc}
			\toprule\rowcolor{gray!20}
			$\tau_{low}$ & $\tau_{high}$ & AP & AP$_{50}$ & AP$_{75}$ & AP$_{S}$ & AP$_{M}$ & AP$_{L}$  \\ \hline\hline
			0.5 & 0.5  &  31.4  &  53.6  & 31.9 & 14.7 & 34.3 & 46.8   \\
			0.4 & 0.6  &  31.9  &  53.9  & 32.3 & 15.4 & 34.8 & 47.7  \\
			0.3 & 0.7  &  32.2  &  54.0  & 33.0 & 15.8 & 34.5 & 48.3  \\ 
			0.2 & 0.8  &  31.8  &  53.6  & 32.8 & 15.5 & 34.1 & 47.2 \\\hline
	\end{tabular}}
	\vspace{-0.7em}
	\caption{Effect of thresholds $\tau_{low}$ and $\tau_{high}$.}
	\label{threshold}
\end{table}

\begin{table}[]
	\centering
	\scalebox{0.8}{
		\begin{tabular}{c|c|cc|ccc}
			\toprule\rowcolor{gray!20}
			$\text{temperature} ({\tau})$ & AP & AP$_{50}$ & AP$_{75}$ & AP$_{S}$ & AP$_{M}$ & AP$_{L}$  \\ \hline\hline
			0.1& 30.9  &  52.5  & 30.9 & 14.5 & 33.0 &  45.0 \\
			0.5& 31.7   & 53.3   & 32.2 & 15.5 & 34.1 & 46.4  \\
			1.0&  32.2  &  54.0  & 33.0 & 15.8 & 34.5 & 48.3  \\
			2.0&  31.9  & 54.1   & 32.1 & 15.6 & 34.0 & 47.7  \\ 
			5.0&  31.6  & 53.8   & 32.5 & 15.6 & 34.2 & 47.5  \\\hline
	\end{tabular}}
	\vspace{-0.7em}
	\caption{Effect of temperature $\tau$.}
	\vspace{-1.0em}
	\label{tau}
\end{table}

\begin{table}[]
	\centering
	\scalebox{0.8}{
		\begin{tabular}{c|c|cc|ccc}
			\toprule\rowcolor{gray!20}
			$\lambda_1$ & AP & AP$_{50}$ & AP$_{75}$ & AP$_{S}$ & AP$_{M}$ & AP$_{L}$  \\ \hline\hline
			0.1& 31.9  &  53.7  & 32.8 & 15.4 & 34.4 & 47.2  \\
			0.3& 32.1   &  54.0  & 33.1 & 15.9 & 34.2 & 47.9  \\
			0.5& 32.2   &  54.0  & 33.0 & 15.8 & 34.5 & 48.3  \\ 
			0.7& 31.4  &  53.6  & 32.2 & 15.2 & 33.9 & 46.9  \\\hline
	\end{tabular}}
	\vspace{-0.7em}
	\caption{Effect of parameter $\lambda_1$.}
	\label{lambda}	
\end{table}

\subsection{Qualitative Results}
Fig.~\ref{vis} shows some qualitative segmentation results of our method and BoxInst~\cite{tian2021boxinst} on COCO $val$ split. On the one hand, our method could better segment foreground instances that heavily tangle with the background or other objects with similar appearances. On the other hand, our method is good at separating overlapping objects of the same semantics and can keep the integrity of objects.

Fig.~\ref{vis2} gives more results of SIM with the ResNet-101-FPN backbone and $3\times$ training schedule. One can see that our method achieves precise predictions around the objects' boundaries.

\subsection{Parameter analysis}
Tab.~\ref{threshold} shows the effects of the two thresholds $\tau_{low}$ and $\tau_{high}$. When we set $\tau_{low}=\tau_{high}=0.5$, all pixels provide supervision, which inevitably introduces much noise. When we set $\tau_{low}=0.2\ \text{and}\ \tau_{high}=0.8$, many pixels are neglected and hence limited supervision is provided and the performance is slightly degraded.

Tab.~\ref{tau} shows the effect of temperature $\tau$. We employ the instance map $M_{\rm I}$ as a weight map to online rectify the semantic map $M_{\rm S}$, where the temperature $\tau$ controls the modulation intensity. When $\tau \to 0$, the modulation intensity increases so that the final pseudo mask $\hat{M}$ will rely more on $M_{\rm S}$. When $\tau \to \infty$, the modulation intensity decreases so that the final pseudo mask $\hat{M}$ will rely more on $M_{\rm I}$.

Tab.~\ref{lambda} shows the segmentation results by using different weights $\lambda_1$ in pseudo loss. One can see that our method is insensitive to this parameter when $0.1<\lambda_1<0.5$.
{\small
	\bibliographystyle{ieee_fullname}
	\bibliography{egbib}
}

\end{document}